# maxVSTAR: Maximally Adaptive Vision-Guided CSI Sensing with Closed-Loop Edge Model Adaptation for Robust Human Activity Recognition

Kexing Liu

*Abstract*— WiFi Channel State Information (CSI)-based human activity recognition (HAR) provides a privacy-preserving, device-free sensing solution for smart environments. However, its deployment on edge devices is severely constrained by domain shift, where recognition performance deteriorates under varying environmental and hardware conditions. This study presents maxVSTAR (maximally adaptive Vision-guided Sensing Technology for Activity Recognition), a closed-loop, vision-guided model adaptation framework that autonomously mitigates domain shift for edge-deployed CSI sensing systems. The proposed system integrates a cross-modal teacher–student architecture, where a high-accuracy YOLO-based vision model serves as a dynamic supervisory signal, delivering real-time activity labels for the CSI data stream. These labels enable autonomous, online fine-tuning of a lightweight CSI-based HAR model, termed Sensing Technology for Activity Recognition (STAR), directly at the edge. This closed-loop retraining mechanism allows STAR to continuously adapt to environmental changes without manual intervention. Extensive experiments demonstrate the effectiveness of maxVSTAR. When deployed on uncalibrated hardware, the baseline STAR model's recognition accuracy declined from 93.52% to 49.14%. Following a single vision-guided adaptation cycle, maxVSTAR restored the accuracy to 81.51%. These results confirm the system's capacity for dynamic, self-supervised model adaptation in privacy-conscious IoT environments, establishing a scalable and practical paradigm for long-term autonomous HAR using CSI sensing at the network edge.

## I. INTRODUCTION

Human Activity Recognition (HAR) has become a foundational capability within intelligent environments, playing a pivotal role in applications spanning smart homes, ambient assisted living, healthcare, and human–machine interaction. In residential settings, HAR enables automated control of home appliances, activity-based service customization, and safety monitoring, thereby improving convenience and occupant well-being [1], [2]. In healthcare contexts, HAR supports the continuous, non-intrusive monitoring of elderly individuals and patients with impaired mobility, facilitating the early detection of falls, abnormal behaviors, and critical health incidents requiring immediate intervention [3]–[5]. Beyond these domains, HAR also contributes to security surveillance [6], human–computer interaction [7], and ubiquitous computing applications [8], [9], where context awareness derived from activity recognition enhances system intelligence and responsiveness. By recognizing daily user activities, systems can develop nuanced behavioral profiles to optimize service delivery — for example, autonomously adjusting environmental parameters, initiating alerts upon detecting unsafe behavior, or providing clinical insights into rehabilitation progress. As IoT-based smart environments continue to expand, advancing HAR systems that are unobtrusive, privacy-preserving, and adaptive becomes increasingly important.

Despite its potential, traditional HAR systems exhibit inherent limitations that restrict their scalability and acceptance in real-world, long-term deployments. Wearable sensor-based approaches, which rely on accelerometers, gyroscopes, or smartwatches to gather motion data, often achieve high recognition accuracy. However, these systems depend on user compliance and may be perceived as intrusive, uncomfortable, or impractical in certain social or clinical scenarios. Users might neglect to wear the devices consistently or refuse their use altogether, limiting the robustness and reliability of such solutions in pervasive monitoring contexts.

In response to these constraints, WiFi Channel State Information (CSI)-based HAR has emerged as a device-free, privacy-conscious alternative that exploits ubiquitous wireless infrastructure for passive activity sensing. CSI provides a fine-grained representation of the wireless signal propagation properties between transmitters and receivers, capturing amplitude and phase information for multiple subcarriers within the wireless spectrum. When a human subject moves within the coverage area, their presence induces fluctuations in CSI values due to signal phenomena such as reflection, refraction, diffraction, and multipath effects. By analyzing these CSI perturbations, activity recognition models can infer and classify the type of human movement taking place.

Modern WiFi standards such as IEEE 802.11n/ac employ Orthogonal Frequency-Division Multiplexing (OFDM), dividing the communication channel into multiple orthogonal subcarriers, each with its own CSI measurement. Consequently, CSI data is typically organized as a complex-valued matrix, with each entry representing the amplitude and phase response of a subcarrier at a given time. This rich, high-dimensional channel information can sensitively capture minute environmental changes, enabling precise recognition of user activities without requiring wearable devices. Notably, many commercial WiFi routers and network interface cards can be configured to export raw CSI values, eliminating the need for additional sensing infrastructure [10]. These characteristics make CSI-based HAR highly attractive for pervasive, low-cost, and privacy-respecting smart environments.

Nonetheless, several formidable challenges hinder the practical deployment of CSI-based HAR systems, particularly on resource-constrained IoT edge devices. Existing approaches often rely on a fixed, predefined activity set, limiting their flexibility and personalization capabilities. Furthermore, the same activity performed in different physical locations, or under varying environmental conditions, may produce substantially different CSI signatures due to dynamic multipath propagation effects. This location dependency complicates the development of robust, generalizable models suitable for wide-area or multi-device IoT deployments.

Another significant issue is the computational burden associated with many state-of-the-art CSI-based HAR models, which are often designed for server-class hardware. These models typically require extensive memory, processing power, and energy resources, making them unsuitable for deployment on lightweight edge devices with constrained resources. While some lightweight HAR models have been proposed for edge AI applications, most represent direct adaptations of traditional methods, lacking architectural optimizations tailored to the unique resource limitations and network dynamics of IoT edge computing environments.

The most critical barrier, however, is the problem of domain shift — a phenomenon where models trained in a specific environment or hardware configuration experience marked performance degradation when applied to new contexts. In the case of CSI-based HAR, minor variations in environmental factors (e.g., room layout, furniture arrangement) or hardware characteristics (e.g., router antenna configuration, device driver versions) can substantially alter CSI patterns. This results in misclassification and unreliable system performance in uncalibrated settings. Traditional remedies, such as collecting labeled data and retraining models for each new deployment, are prohibitively labor-intensive and impractical at scale, particularly in dynamic or mobile IoT environments.

To address these challenges and enable reliable, long-term CSI-based HAR in practical IoT deployments, this paper proposes maxVSTAR (maximally adaptive Vision-guided Sensing Technology for Activity Recognition) — a novel closed-loop, cross-modal framework designed for edge-deployed, privacy-preserving HAR systems. The core innovation of maxVSTAR lies in its use of a teacher–student architecture, wherein a high-precision vision-based activity recognition model, built upon the YOLO object detection framework, serves as an adaptive supervisory signal. This YOLO-based vision model generates real-time ground truth labels for the CSI data stream. These labels enable the lightweight CSI-based Sensing Technology for Activity Recognition (STAR) model to perform continuous, online fine-tuning directly on edge devices.

By operating entirely at the network edge, maxVSTAR eliminates the need to transmit CSI or video data to remote servers, ensuring privacy and reducing communication overhead. Its closed-loop retraining process allows the STAR model to autonomously adapt to new environmental conditions and hardware configurations, overcoming domain shift without manual recalibration. This design ensures that HAR systems maintain high recognition accuracy and operational stability in dynamic, real-world IoT environments while conforming to stringent edge device resource constraints.

Through extensive experimental validation, this study demonstrates that maxVSTAR effectively mitigates domain shift in CSI-based HAR applications, recovering significant performance losses encountered in new deployment scenarios. The framework offers a scalable, efficient, and privacy-respecting paradigm for long-term, autonomous human activity recognition in pervasive smart environments.

The remainder of this paper is organized as follows: Section II reviews related work in Wi-Fi-based HAR and embedded edge sensing systems. Section III details the design of the proposed methodology, including data acquisition, signal processing, and adaptive mechanisms including synchronization and online learning in maxVSTAR. Section IV presents the experimental setup, covering data collection environments, model deployment configurations, and evaluation metrics. Section V concludes with a summary of findings and directions for future research.

## II. RELATED WORKS

In recent years, human activity recognition (HAR) using WiFi signals has attracted considerable research interest as a non-contact, device-free sensing modality for smart environments [1]. The core principle involves leveraging the influence of human movements on wireless signal propagation. As individuals move within a WiFi-covered area, their bodies perturb the signal through reflection, diffraction, and scattering, leading to measurable variations in signal characteristics at the receiver [1], [2]. Early studies predominantly relied on the Received Signal Strength Indicator (RSSI) for HAR [3]. However, RSSI provides only coarse-grained signal strength information, lacking the spatial resolution necessary for fine-grained activity recognition and being highly susceptible to multipath interference and environmental noise [2], [3].

The advent of more advanced wireless communication standards has enabled the use of Channel State Information (CSI) for HAR [4]–[13]. CSI captures detailed amplitude and phase information across multiple Orthogonal Frequency-Division Multiplexing (OFDM) subcarriers, offering a finer characterization of wireless channel conditions. Compared to RSSI, CSI is more sensitive to subtle environmental variations and minor human motions, including activities as fine-grained as respiration [14], [15]. Recent literature has explored various CSI feature extraction techniques, including statistical descriptors, time–frequency representations, and deep learning-based approaches, alongside classifiers such as Support Vector Machines (SVM), K-Nearest Neighbors (KNN), Convolutional

Neural Networks (CNN), Recurrent Neural Networks (RNN), and Transformer models [16]. WiFi CSI-based HAR has demonstrated potential in diverse applications, including activity detection, presence sensing, fall detection, and gesture recognition [15], [17]–[21].

The integration of deep learning has notably advanced CSI-based HAR in recent years. CNNs have proven effective for extracting spatial and temporal features from CSI matrices, while RNNs, particularly Long Short-Term Memory (LSTM) networks, have excelled in modeling the time-series characteristics of CSI data. More recently, Transformer architectures have shown promise due to their ability to capture long-range dependencies in sequential data [22]–[31]. For example, Zhang et al. [23] proposed a WiFi CSI-based HAR workflow, comparing InceptionTime and LSTM classifiers while analyzing hardware-related signal variability. Hnoohom et al. [24] introduced a deep residual network architecture for CSI-based HAR and benchmarked multiple deep models, including CNNs, LSTMs, GRUs, and bidirectional variants, achieving a recognition accuracy of 98.60%, outperforming previous benchmarks by 3.60%.

Several dedicated CSI-HAR architectures have been proposed, such as DF-CNN [29], SLNet [28], RF-Net [27], and WiFlexFormer [26], each tailored to the unique signal characteristics of CSI data and aiming to improve recognition accuracy and computational efficiency. Collectively, these works have positioned deep learning as the mainstream approach for CSI-based HAR, with ongoing efforts focused on architectural optimization and model robustness.

Simultaneously, the growing demand for low-latency, privacy-preserving HAR systems has catalyzed a migration of AI workloads from centralized cloud infrastructures to edge computing platforms. Edge deployment is particularly critical in safety-critical HAR use cases, such as fall detection and security surveillance, where response latency must be minimized. However, deploying deep learning models on resource-constrained edge hardware—such as NVIDIA Jetson modules [32], Google Coral Edge TPU [33], Raspberry Pi [34], Qualcomm SoCs [35], and Thundercomm's RUBIK Pi3 [36]—presents significant challenges due to limited processing, memory, and power budgets. Recent edge SoCs integrating AI accelerators, including Rockchip's RK3588 and RV1126 [37], [38], and Hailo's Hailo-8 [39], offer enhanced compute capabilities, yet HAR models must still be carefully optimized to meet these resource constraints.

Model compression techniques have therefore become a core strategy for deploying AI models on edge platforms. Francy and Singh [40] comprehensively reviewed model compression strategies, including pruning [41], quantization [42], knowledge distillation [43], and low-rank factorization [44], each designed to reduce model size and computational complexity while maintaining acceptable accuracy. Additionally, Neural Architecture Search (NAS) has been employed to discover efficient model architectures tailored to specific hardware and application constraints [45]. Modern edge devices also integrate hardware AI accelerators such as GPUs, NPUs, DSPs, FPGAs, and ASICs, necessitating models designed to fully exploit these specialized resources [46].

Another persistent challenge in deep learning-based HAR is the labor-intensive process of creating large, high-quality annotated datasets, particularly for non-visual sensing modalities such as CSI. While extensive computer vision datasets exist, CSI datasets are limited, and manual annotation remains time-consuming and costly [47]. To address this, researchers have explored using real-time object detection frameworks like YOLO for rapid annotation of video data. Mokdad et al. [48] evaluated YOLO-based automatic video annotation systems, emphasizing their value in accelerating dataset generation. Additionally, semi-automatic annotation methods, including human-in-the-loop workflows [49], have been proposed to reduce annotation effort. These techniques suggest a broader cross-modal strategy, where a reliable "teacher" modality such as vision can provide real-time ground truth labels for a lower-interpretability "student" modality like CSI.

Despite these advances, a critical gap persists. While prior works have focused on improving model architectures, enhancing feature extraction, and addressing computational constraints, most studies treat domain shift—the performance degradation encountered when models are transferred to new environments—as a static problem. Existing approaches attempt to mitigate domain shift by creating inherently more robust models or retraining models offline with additional data. However, none have proposed a practical, closed-loop, self-adapting framework capable of performing on-device, autonomous model fine-tuning using real-time supervisory labels generated by a cross-modal "teacher" model.

To the best of our knowledge, no existing work has validated an integrated system that employs a high-precision vision-based activity recognizer to provide continuous, real-time supervision for a CSI-based HAR model, enabling dynamic online adaptation directly on edge devices. This capability is especially crucial for long-term, privacy-sensitive deployments in IoT environments, where environmental dynamics and hardware variations are inevitable, and cloud-based recalibration is impractical due to privacy, latency, and bandwidth constraints.

Our proposed maxVSTAR framework directly addresses this unmet need. By combining a cross-modal teacher–student paradigm with a closed-loop, on-device model update mechanism, maxVSTAR achieves real-time, autonomous model adaptation for CSI-based HAR at the edge. Its unique design ensures long-term operational robustness without cloud intervention, overcoming domain shift while preserving data privacy and meeting edge device resource limitations. This work establishes a novel, scalable paradigm for self-adapting, cross-modal IoT sensing systems—a capability not currently achievable with existing state-of-the-art solutions.

III. METHODOLOGY

*A. Overview of the maxVSTAR Framework*

This study proposes maxVSTAR (maximally adaptive

Vision-guided Sensing Technology for Activity Recognition) — a novel framework designed to address a critical bottleneck in real-world WiFi Channel State Information (CSI)-based Human Activity Recognition (HAR): domain shift. While existing static models can achieve high recognition accuracy under controlled conditions, their performance typically degrades in new or dynamic environments due to environmental fluctuations and hardware variations.

WiFi CSI is represented as a complex-valued tensor:

$$H(t) \in \mathbb{C}^{N_t \times N_r \times K}$$

where $N_t$ and $N_r$ enote the number of transmitting and receiving antennas respectively, and $K$ is the number of subcarriers. The temporal sequence of CSI measurements

$$\mathcal{H} = \{H(t_1), H(t_2), \dots, H(t_T)\}$$

serves as the input to the HAR model, whose goal is to classify each time point into one of $C$ predefined human activity categories, such that

$$y(t) \in \{1, 2, \dots, C\}$$

To overcome performance degradation caused by domain shift, maxVSTAR introduces a closed-loop, self-adaptive system that enables CSI-based HAR models to autonomously recalibrate and update on edge devices without cloud interaction. The adaptation relies on cross-modal supervision where image-based activity detection acts as a reliable ground-truth source for CSI-based model retraining.

To preserve privacy and mitigate user discomfort associated with continuous visual monitoring, image-based detection is only activated during the model adaptation phase. Prominent notifications are issued to monitored subjects during this brief vision-assisted retraining, after which the system reverts to CSI-only inference. The overall architecture of maxVSTAR is depicted in Fig. 1.

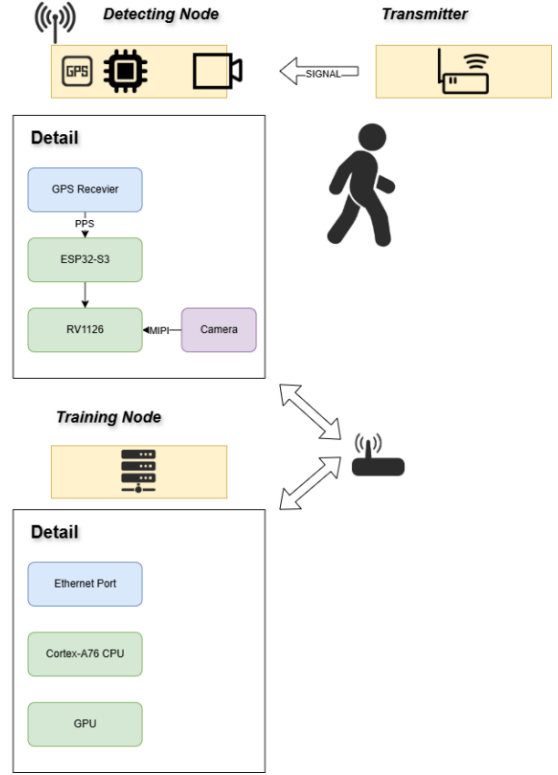

**Fig. 1.** Overall system architecture of maxVSTAR.

maxVSTAR follows a distributed edge AI framework separating real-time inference and model adaptation tasks. The system comprises two core components:
1. Detection Nodes: Lightweight inference units running the baseline STAR model deployed at the edge [50].
2. Training Node (Edge Server): A more powerful edge node responsible for model fine-tuning and update distribution.

The inference model operates by processing CSI sequences into feature vectors $X(t) \in \mathbb{R}^d$ subsequently classified by a sequence model. The training node operates between typical inference edge devices and server-class GPUs in terms of compute capability. Its hardware specifications are listed in Table I.

TABLE I. Hardware specifications of maxVSTAR training node

| Component | Specification |
| --- | --- |
| CPU | Eight-core ARM Cortex-A76 |
| RAM | 32 GB |
| Storage | 128 GB SSD |
| Network | 2.5 Gbps |
| OS | Yocto Linux |
| ML Framework | PyTorch |

A system block diagram of the training node is provided in Fig. 2.

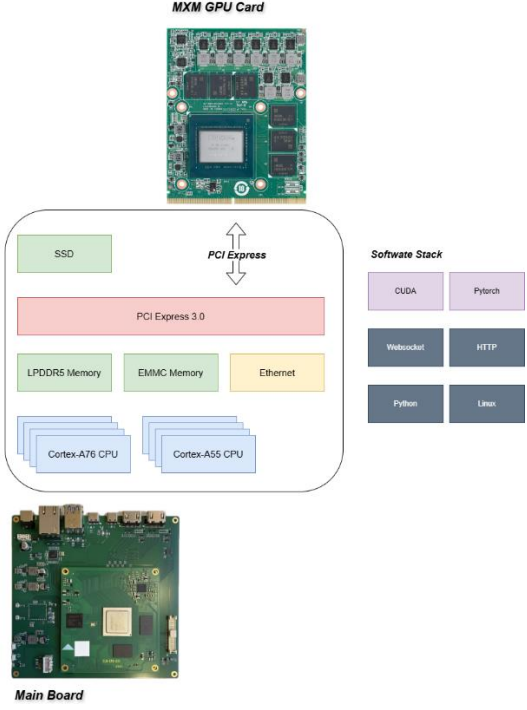

**Fig. 2.** Block diagram and physical appearance of the maxVSTAR training node.

*B. The Foundational "Student" Model: STAR*

The core inference model within the framework is STAR (Sensing Technology for Activity Recognition) — a lightweight, three-layer Gated Recurrent Unit (GRU) network tailored for real-time CSI-based HAR on resource-constrained devices. The model ingests sequential CSI amplitude and/or phase data, formatted as

$$\mathcal{X} = \{x(t_1), x(t_2), \ldots, x(t_T)\}, x(t_i) \in \mathbb{R}^d$$

where $d$ is the dimensionality of the flattened CSI feature vector per timestamp, constructed after pre-processing operations like Short-Time Fourier Transform (STFT) and normalization:

$$x(t) = Norm(|\mathcal{F}(H(t))|)$$

with $\mathcal{F}$ representing the Fourier transform operator. The GRU network applies gated operations at each time step $t$ according to:

- Update gate:

$$z_t = \sigma(W_z x_t + U_z h_{t-1} + b_z)$$

- Reset gate:

$$r_t = \sigma(W_r x_t + U_r h_{t-1} + b_r)$$

- Candidate hidden state:

$$\widetilde{h_t} = tanh(W_h x_t + U_h(r_t \odot h_{t-1}) + b_h)$$

- Final hidden state:

$$h_t = (1 - z_t) \odot h_{t-1} + z_t \odot \widetilde{h_t}$$

Here, $W_*$, $U_*$, $b_*$, $b_*$ are learnable parameters, $\sigma(\cdot)$ is the sigmoid activation, and $\odot$ denotes element-wise multiplication. The final output sequence is passed through a softmax classifier:

$$\hat{y}_t = argmax(Softmax(W_o h_t + b_o))$$

Under ideal, calibrated conditions, STAR achieves a baseline classification accuracy of 93.52% [50]. Its model architecture and optimization strategies for the Rockchip RV1126 platform have been detailed in prior work. Within maxVSTAR, STAR serves as the "student" model, receiving adaptive weight updates from the training node during domain shifts to maintain recognition robustness while minimizing edge device compute overhead.

*C. The Vision "Teacher" Model: Enhanced YOLO Architecture*

Given the abstract and difficult-to-interpret nature of CSI data, reliable supervisory labels are critical for successful adaptive learning. The feasibility of the closed-loop framework hinges on the accuracy of the vision-based "teacher" model; erroneous labels would propagate errors into the CSI model during adaptation.

To ensure high-fidelity supervisory signals, we developed an improved object detection model based on YOLOv8, incorporating a lightweight attention module (iRMB) integrated with the C2f module, forming the novel C2f_iRMB block (see Fig. 3).

Formally, given an input image tensor $I \in \mathbb{R}^{H \times W \times 3}$, the YOLO detection pipeline computes a dense prediction map $\mathcal{P} \in \mathbb{R}^{H' \times W' \times B(C+5)}$. where $B$ is the number of anchor boxes per grid cell, and each anchor predicts:

- 4 bounding box coordinates,
- 1 objectness score, and
- $C$ class probabilities.

For each grid cell, detection confidence is computed as:

$$Conf(i,j) = P_{obj}(i,j) \times IoU_{pred,gt}$$

The proposed C2f_iRMB module integrates an efficient residual attention mechanism into the C2f bottleneck structure. Mathematically, the iRMB attention mechanism computes:

- Channel-wise attention:

$$s_c = Sigmoid\left(FC_2\left(\delta\left(FC_1(GAP(F))\right)\right)\right)$$

where GAP is global average pooling, $\delta$ is ReLU activation,

and $FC_*$ are fully connected layers.
- Spatial-wise attention (if integrated):

$$\mathbb{s}_s = \sigma\left(Conv_{7\times 7}\left(Concat(AvgPool(F), MaxPool(F))\right)\right)$$

The attended feature map is then computed as:

$$F_{att} = \mathbb{s}_c \odot F$$

This enhancement improves detection performance in complex scenes involving occlusions and clutter, ensuring accurate real-time pose detection. Experiments indicate that YOLOv8 augmented with C2f_iRMB achieved a 21% accuracy improvement on the internal dataset compared to its vanilla implementation.

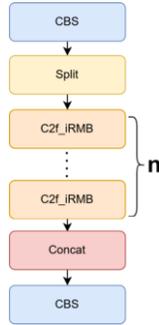

**Fig. 3.** Architecture of the proposed C2f_iRMB module integrated into YOLOv8.

*D. Core Adaptive Mechanism: Synchronization and Closed-Loop Online Learning*

**1) Multi-Modal Time Synchronization**

Precise synchronization between the CSI data stream and vision-based labels is fundamental for reliable model adaptation in maxVSTAR. The system implements a dual-layer synchronization strategy combining hardware-level and software-level mechanisms to ensure that each CSI sample $x(t)$ is ccurately paired with the corresponding visual label $y_{vision}(t)$.

**Hardware-Level Synchronization**:
A GPS receiver supplies a high-precision Pulse-Per-Second (PPS) signal providing the global time base. Each CSI packet is assigned a timestamp:

$$t_{CSI}^{(t)} = T_{PPS} + \Delta t^{(i)}$$

where $T_{PPS}$ is the latest PPS reference and $\Delta t^{(i)}$ is the relative offset measured by a synchronized high-resolution hardware counter. Simultaneously, each frame from the MIPI image sensor is timestamped via a deterministic hardware trigger (see Fig. 4):

$$t_{img}^{(j)} = T_{PPS} + \delta t^{(j)}$$

ensuring absolute alignment:

$$\left|t_{CSI}^{(i)} - t_{img}^{(j)}\right| \leq \varepsilon_{sync}$$

where $\varepsilon_{sync}$ is a nanosecond-scale threshold.

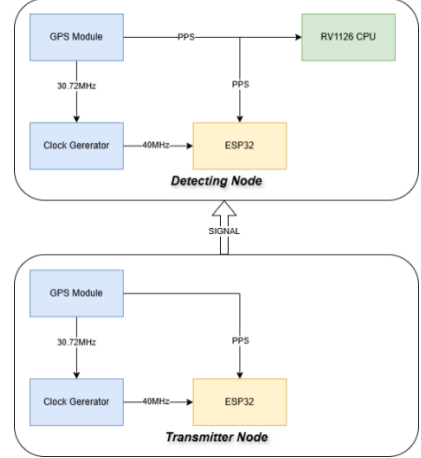

**Fig. 4.** Hardware clock synchronization principle for CSI and image streams.

**Software-Level Optimization**:
To further reduce latency and jitter, a real-time kernel context elevates the priority of CSI and image acquisition threads. Data packets are written to a ring buffer located in physical address space:

$$Buffer_{phy} = \{\mathcal{X}_k, \mathcal{T}_k\}_{k=1}^N$$

using direct memory mapping via mmap() to bypass the MMU, which eliminates costly user-kernel transitions:

$$Latency_{copy} \to 0$$

This achieves ultra-low-latency, high-throughput, multi-modal data acquisition and synchronization, ensuring minimal timestamp deviation at inference-time fusion.

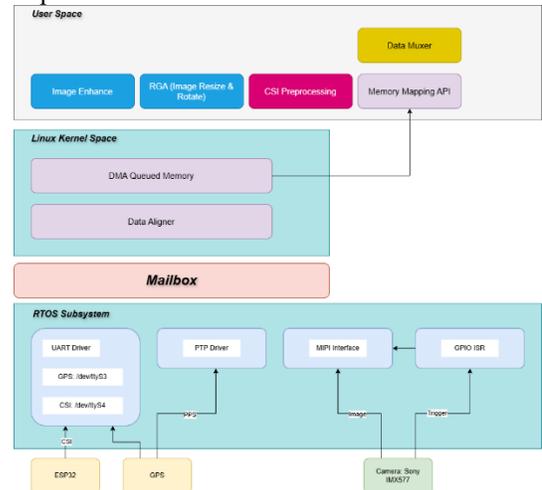

**Fig. 5.** Optimized acquisition and memory-mapped storage dataflow.

### 2) Closed-Loop Fine-Tuning Workflow

The closed-loop adaptation mechanism of maxVSTAR implements a five-step workflow, designed to automatically detect model degradation and retrain the STAR model on newly labeled data with minimal human intervention. Mathematically, this process can be described as follows:

At time $t$, a detection node triggers an update request if its activity recognition confidence drops below a threshold:

$$IF\ \mathbb{E}_{t \in T}[Conf_{CSI}(t)],\ trigger\ update$$

The camera activates, generating a live video stream $\mathcal{T}_t$. The improved YOLOv8 C2f_iRMB model processes the image frames to produce activity labels:

$$y_{vision}(t) = YOLOv8_{C2f\_iRMB}(\mathcal{T}_t)$$

Simultaneously, synchronized CSI data $x(t)$ is acquired. After ensuring timestamp alignment:

$$|t_{CSI}(t) - t_{vision}(t)| \leq \varepsilon_{sync}$$

the activity label is assigned to the CSI sample:

$$x(t) \rightarrow y_{vision}(t)$$

Labeled data pairs $\{x(t), y_{vision}(t)\}$ are transmitted to the training node, where the STAR model is fine-tuned using a supervised loss function:

$$\mathcal{L}_{CSI} = -\sum_{c=1}^{C} \mathbf{1}_{[y_{vision}=c]} \cdot \log p_c(t)$$

where $p_c(t)$ is the predicted probability for class $c$ at time $t$, obtained via STAR's softmax layer. After optimization via backpropagation-through-time (BPTT) and Adam optimizer:

$$\theta_{CSI}^{(t+1)} \leftarrow \theta_{CSI}^{(t)} - \eta \frac{\partial \mathcal{L}_{CSI}}{\partial \theta_{CSI}}$$

the updated model parameters $\theta_{CSI}^{(t+1)}$ are distributed to all online detection nodes. Following this process, the vision module is deactivated to preserve privacy, and inference continues solely via CSI-based sensing using the updated model. The closed-loop operation of maxVSTAR proceeds through the following five steps (illustrated in Fig. 6):

1. The detection node initiates a model update request.
2. The detection node activates its camera to continuously collect image data and uses a pre-trained YOLO model to determine poses.
3. Simultaneously, CSI data is collected. After ensuring time alignment, corresponding labels are added to the CSI data.
4. The labeled CSI data is packaged and uploaded to the training node.
5. After the training node completes the model update, it distributes the new weights to all online detection nodes, completing the adaptive fine-tuning of the model.

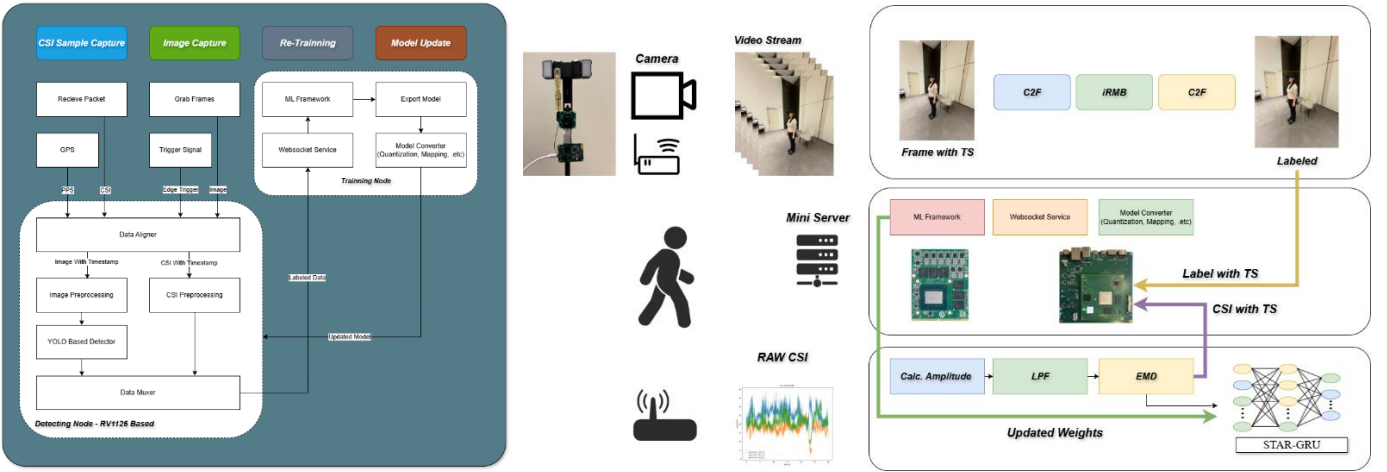

**Fig. 6.** Closed-loop workflow of the maxVSTAR framework.

*E. System Deployment and Co-Optimization on the Edge*

maxVSTAR's deployment strategy exploits the heterogeneous compute architecture of the Rockchip RV1126 platform, integrating a 2.0 TOPS Neural Processing Unit (NPU), Raster Graphics Accelerator (RGA), and ARM Cortex-A55/A76 CPU cores to achieve efficient, low-latency multi-modal inference and adaptation at the edge.

**YOLO Model Deployment**:

To accommodate 4K video streams, image downscaling is offloaded to the RGA, while an initial person-detection pre-filter reduces computational load by focusing on regions of interest. The pre-trained YOLOv8 C2f_iRMB model is exported in ONNX format and deployed on the RV1126's NPU for hardware-accelerated object detection. Each input video frame $\mathcal{T}_t$ of resolution $W \times H$ undergoes an initial image downscaling operation:

$$\mathcal{T}_t^{scaled} = RGA_{down}(\mathcal{T}_t)$$

where

$$(W_{scaled}, H_{scaled}) = \alpha \times (W, H)$$

with scaling factor $\alpha \in (0,1]$ computed based on available NPU memory capacity and inference time constraints. A lightweight person-detection pre-filter computes a binary mask:

$$M_{person}(x,y) = \begin{cases} 1, & \text{if person detected at } (x,y) \\ 0, & \text{otherwise} \end{cases}$$

which restricts the subsequent YOLO object detection to regions of interest (ROI), reducing computational complexity from $\mathcal{O}(W,H)$ to $\mathcal{O}(W',H')$, where $(W',H') \ll (W,H)$. The YOLO detection head produces a set of bounding boxes and class probabilities:

$$\mathcal{B} = \left\{ \left(x_{max}^{(i)}, x_{min}^{(i)}, y_{max}^{(i)}, y_{max}^{(i)}, p_c^{(i)}\right) \right\}_{i=1}^{N_{det}}$$

where $N_{det}$ is the number of detections in a given frame.

**CSI Model Deployment**:

The lightweight GRU-based STAR model is also mapped onto the NPU for inference. Each CSI sample stream $\mathcal{X} = \{x_1, x_2, \ldots, x_T\}$ undergoes pre-processing steps including Fourier Transform for denoising:

$$\hat{x}_t = \mathcal{F}(x_t)$$

and min-max normalization:

$$\hat{x}_t = \frac{-min(\hat{x}_t)}{max(\hat{x}_t) - min(\hat{x}_t)}$$

To expedite these operations, vectorized ARM NEON SIMD instructions are used, accelerating both spectral computations and element-wise transformations. The resulting normalized sequence $\{\tilde{x}_1, \ldots, \tilde{x}_T\}$ is fed into the STAR network, where the hidden state update at time $t$ for layer $l$ as follows:

$$\begin{aligned} h_t^{(l)} = &(1 - z_t^{(l)}) \odot h_{t-1}^{(l)} \\ &+ z_t^{(l)} \odot tanh\left(W_h^{(l)} x_t + U_h^{(l)}\left(r_t^{(l)} \odot h_{t-1}^{(l)}\right)\right. \\ &\left. + b_h^{(l)}\right) \end{aligned}$$

with standard GRU gate formulations:

$$z_t^{(l)} = \sigma\left(W_z^{(l)} x_t + U_z^{(l)} h_{t-1}^{(l)} + b_z^{(l)}\right)$$
$$r_t^{(l)} = \sigma\left(W_r^{(l)} x_t + U_r^{(l)} h_{t-1}^{(l)} + b_r^{(l)}\right)$$

ensuring efficient temporal sequence modeling.

**Memory and Dataflow Optimization**:

All inference operations for both CSI and vision streams are executed in physical memory regions accessed via memory mapping:

$$\mathcal{M}_{phys} = mmap(0, size, PROT^{read}|PROT^{rewrite}, MAP^{SHARED}, fd, 0)$$

This configuration bypasses the Memory Management Unit (MMU), eliminating kernel-to-user data copying:

$$Latency_{copy} \approx 0$$

and reducing total inference time per frame:

$$T_{inference} = T_{CSI\_proc} + T_{YOLO\_inference} + T_{memory\_mapping}$$

with typical values satisfying $T_{inference} < 50ms$ to meet real-time HAR system constraints.

The maxVSTAR system partitions tasks according to hardware specialization, minimizing inter-component data transfer:

- YOLO object detection → NPU
- Image pre-scaling and ROI extraction → RGA
- CSI spectral preprocessing → NEON SIMD CPU
- CSI inference (STAR GRU layers) → NPU
- Model fine-tuning (when needed) → Edge server CPU/GPU

The task allocation and hardware mapping strategy are illustrated in Fig. 7.

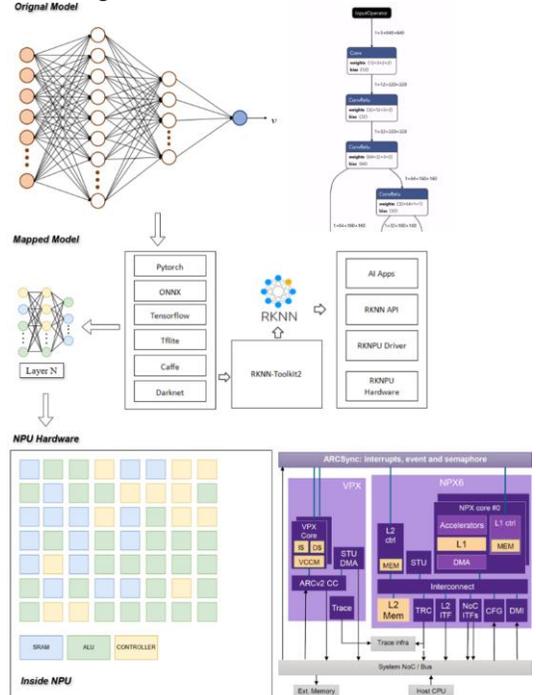

**Fig. 7.** Hardware resource mapping for CSI inference, vision processing, and adaptation on the RV1126.

## IV. EXPERIMENTAL SETUP AND VALIDATION

### A. Experimental Environment

To rigorously evaluate the proposed maxVSTAR framework under realistic IoT edge computing constraints, we constructed a dedicated experimental testbed composed of heterogeneous hardware and software components. The detection node was built upon a Rockchip RV1126 system-on-chip (SoC) platform, integrating a 2.0 TOPS Neural Processing Unit (NPU) and a Sony IMX577 4K MIPI camera for visual sensing. WiFi Channel State Information (CSI) data was concurrently acquired via an onboard ESP32-S3 wireless module interfaced through a high-speed SPI connection. To handle model fine-tuning and closed-loop adaptation, a higher-performance edge training node was configured using an Intel Core i7-14700 CPU, 64 GB of memory, and an NVIDIA RTX 4060Ti GPU. All system components operated under a custom-built Yocto/Buildroot Linux distribution, ensuring a deterministic runtime environment and full control over kernel-level real-time task scheduling.

Deep learning models, including the baseline STAR network and the YOLO-based vision model, were implemented using PyTorch 2.0.1. To precisely simulate hardware-induced domain shifts, an IQXEL-M professional wireless signal testing platform was employed. This setup facilitated controlled modifications of channel impairments, including adjustable attenuation, antenna pattern alterations, and multipath reflectors, thereby enabling reproducible domain shift scenarios for systematic evaluation.

Two datasets were meticulously prepared to support model training, validation, and adaptive fine-tuning within the maxVSTAR framework. The first, a WiFi Channel State Information (CSI) dataset, consists of 200,000 CSI data samples collected at a sampling rate of 100 Hz within a controlled indoor laboratory environment measuring 5.5 m × 6.5 m. The environment was deliberately designed to introduce moderate variability in multipath propagation while avoiding uncontrolled interference, with typical WiFi signal-to-noise ratios (SNR) ranging from 35 to 48 dB. This dataset covers seven distinct human activity classes — lying down, falling, walking, picking up, running, sitting down, and standing up — along with a null state representing the absence of human presence. Data were collected using a Rockchip RV1126 edge node paired with an ESP32-S3 module for CSI signal acquisition, with synchronized timestamping provided via a hardware GPS Pulse-Per-Second (PPS) signal to ensure precise alignment of CSI data with ground-truth labels. Each activity was performed by five subjects with varied physical characteristics, each repeating the motions in randomized sequences to ensure intra-class variability and enhance model generalizability.

The CSI dataset was partitioned into training, validation, and test subsets at a fixed ratio of 8:1:1, ensuring balanced distribution of activity classes across splits while preserving subject independence in the test set to prevent identity bias in evaluation.

To supervise the CSI model's online adaptive fine-tuning, a self-annotated visual dataset containing 8,812 images was independently created. This dataset captured the same seven activity classes under diverse environmental lighting, occlusion, and perspective conditions. Image data were collected using a Sony IMX577 4K camera module integrated into the RV1126 detection node, ensuring identical sensor perspectives for both CSI and vision modalities. Additional images were sourced from public datasets and web-scraped repositories using automated crawlers, followed by rigorous manual curation and annotation with activity labels. Due to the significant workload involved in verifying and cleaning this multimodal image collection, assistance from a multimodal LLM tool (LLaMA 2) was employed for automated image clustering and initial pose classification prior to final human verification.

The visual dataset underwent extensive preprocessing, including duplicate removal, label consistency validation, and data augmentation techniques such as random rotation (±15°), scaling (0.8×–1.2×), horizontal flipping, brightness adjustment (±20%), and color jittering to simulate realistic environmental variations and enhance the robustness of the YOLO detection model.

The spatial arrangement of the CSI transceivers, the RV1126 vision detection node, and the data collection proximity zones is schematically illustrated in Figure 8, detailing the relative positions of transmitters, receivers, and monitored activity zones within the testbed environment.

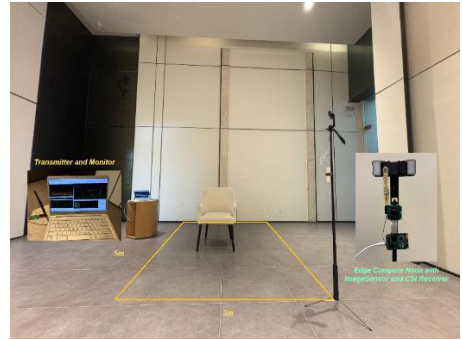

**Fig. 8.** Experimental data acquisition setup and spatial layout.

The figure illustrates the controlled indoor environment used for CSI and visual data collection. A 5 m × 3 m rectangular activity zone was delineated on the floor, within which human subjects performed seven predefined activity classes for CSI-based HAR model training and evaluation. A chair was positioned at the center of the rectangle to serve as a prop for seated and transitional actions. At two diagonally opposing corners of the rectangle, a laptop-based WiFi CSI transmitter and a tripod-mounted Rockchip RV1126 edge compute node equipped with a 4K Sony IMX577 camera and an ESP32-S3 CSI receiver were installed. This diagonal transceiver arrangement was selected to maximize multipath diversity and capture signal variations induced by human movement. The CSI acquisition node and the vision module were synchronized via a GPS PPS signal to ensure precise temporal alignment between the wireless signal packets and the corresponding image frames. This configured layout ensured consistent and repeatable data acquisition conditions while preserving sufficient environmental variability to validate the domain shift resilience of the maxVSTAR framework.

*B. Validation Protocol*

The experimental validation was carefully designed to evaluate both the baseline performance of the STAR model and the self-adaptive capabilities of the proposed maxVSTAR system under controlled domain shift conditions. Initially, the baseline STAR model was trained and tested on the fully calibrated hardware environment to establish reference classification accuracy. Subsequently, domain shift conditions were introduced progressively by altering antenna orientations, introducing multipath reflectors, and applying calibrated signal attenuation using the IQXEL-M test system. This systematic manipulation allowed for quantifiable degradation of HAR model performance, simulating real-world deployment variability.

To verify the effectiveness of the vision-guided closed-loop adaptation mechanism, the maxVSTAR system was activated following each induced domain shift. During this process, the detection node simultaneously collected CSI data and synchronized visual streams, with the latter processed by the enhanced YOLO-based "teacher" model to generate reliable activity labels in real time. The labeled CSI data was then transferred to the edge training node, where the STAR model underwent on-device fine-tuning for 50 epochs using a fixed learning rate of 0.0001. Once adaptation was complete, the updated model was redistributed to the detection node, and performance was re-evaluated under the same domain shift conditions.

Recognition accuracy was calculated as the percentage of correctly predicted activity labels over the total number of test samples. Mean Average Precision (mAP) was used to assess the detection performance of the YOLO-based vision model at Intersection-over-Union (IoU) thresholds of 0.5 and 0.75. All experiments were independently repeated five times, and results were reported as mean values accompanied by standard deviations to account for random variation.

This integrated experimental framework was explicitly designed to capture the challenges inherent to IoT edge-based HAR applications, including resource-constrained real-time inference, strict privacy preservation by avoiding cloud interaction, and resilience against dynamic environmental and hardware-induced domain shifts. By covering these operational requirements, the evaluation substantiates both the technical effectiveness and the practical deployability of the maxVSTAR framework in real-world smart environment scenarios.

V. EXPERIMENTAL RESULTS AND PERFORMANCE VALIDATION

To comprehensively evaluate the effectiveness of the proposed maxVSTAR framework, we conducted a series of controlled experiments addressing two core objectives: validating the detection performance of the improved YOLO-based "teacher" model and assessing the adaptive capability of the closed-loop maxVSTAR system under severe domain shift scenarios. The results substantiate both the technical validity and practical relevance of the proposed approach.

*A. Validation of the Improved YOLO "Teacher" Model*

The performance of the YOLOv8n-C2f_iRMB model, central to the closed-loop adaptation mechanism, was first assessed through comparative analysis against the baseline YOLOv8n configuration. Both models were trained on the curated visual pose dataset, with convergence and stability observed throughout the training process. To quantify detection performance, we computed precision-recall (PR) curves, presented in Figures 9 and 10 for the baseline and improved models, respectively. The results clearly illustrate a consistently superior PR curve for the YOLOv8n-C2f_iRMB model, indicating improved classification performance across the full range of recall thresholds.

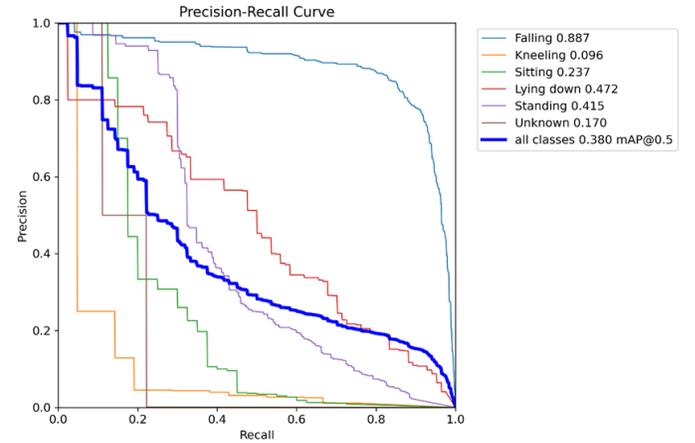

**Fig. 9.** Precision-Recall (PR) curve of the baseline YOLOv8n model on the internal visual pose dataset. The curve reflects the detection trade-off between precision and recall across varying thresholds, illustrating the limitations of the original configuration in complex activity detection scenarios.

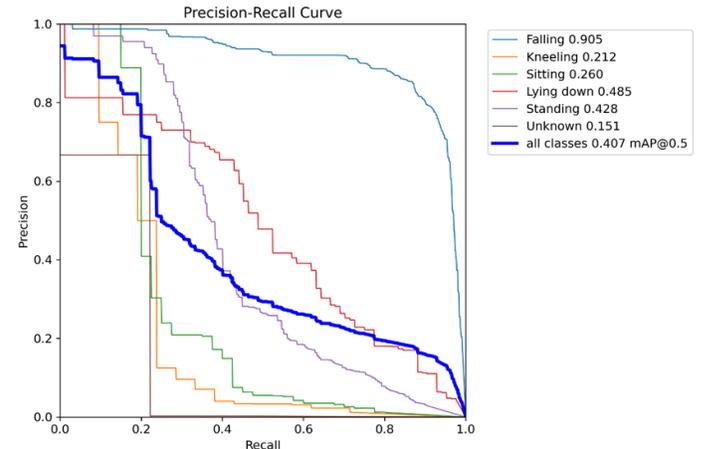

**Fig. 10.** Precision-Recall (PR) curve of the proposed YOLOv8n-C2f_iRMB model on the same dataset. The improved model demonstrates consistently higher precision at all recall levels, confirming the efficacy of the C2f_iRMB module in enhancing detection accuracy and reducing false positives in occluded and cluttered environments.

In Table II, the improved model achieved a substantial increase in mean Average Precision (mAP) at IoU 0.5, rising from 37.9% to 40.7%. More critically, precision exhibited a remarkable enhancement of 21.9 percentage points, improving from 48.4% to 70.3%. This considerable increase demonstrates a substantial reduction in false positives, a crucial characteristic for generating reliable supervisory signals in the adaptive framework. While recall experienced a modest decline of 1.1%, the precision gain significantly outweighs this trade-off in the context of cross-modal supervision, where avoiding noisy or incorrect labels is of paramount importance. Computational complexity, reflected by an increase in GFLOPS from 8.1 to 11.2, remained within acceptable operational limits for the designated edge hardware platform. These findings confirm that the C2f_iRMB module enhances the model's discriminative capacity for fine-grained activity recognition tasks, validating its role as a high-quality "teacher" model within the maxVSTAR framework.

TABLE II. RESULTS OF THE ABLATION STUDY ON THE C2F_IRMB MODULE

| Model | mAP (0.5)/% | mAP (0.5:0.95)/% | P/% | R/% | GFLOPS |
|---|---|---|---|---|---|
| YOLOv8n | 37.9 | 30.3 | 48.4 | 36.7 | 8.1 |
| YOLOv8n-C2f_iRMB | 40.7 | 32.9 | 70.3 | 35.6 | 11.2 |

*B. Validation of the maxVSTAR Adaptive Framework*

Following the validation of the "teacher" model, the study proceeded to evaluate the core functionality of the maxVSTAR closed-loop system under operational conditions that simulate severe domain shift. To establish a baseline, the pretrained STAR model was tested directly on an uncalibrated hardware configuration. The results, detailed in Table III, confirmed a substantial deterioration in classification accuracy. The average recognition rate across all activity classes declined precipitously from 93.52% in calibrated settings to 49.14% under uncalibrated conditions, with particularly poor performance observed in the "stand up" and "sit down" activities, at 17.43% and 26.90% accuracy, respectively. This performance collapse highlights the fragility of static CSI-based HAR models in real-world, dynamically varying environments and underscores the necessity of adaptive mechanisms to maintain operational viability.

TABLE III. STAR MODEL CLASSIFICATION ACCURACY ON UNCALIBRATED DEVICES

| Class of activity | Accuracy |
|---|---|
| lie down | 36.22% |
| fall | 55.44% |
| walk | 56.10% |
| pickup | 44.15% |
| run | 27.99% |
| sit down | 26.90% |
| stand up | 17.43% |
| Have a person or No person | 79.23% |

Upon confirming the severity of the domain shift, the maxVSTAR workflow was activated. Uncalibrated CSI data was collected and simultaneously annotated in real-time via the improved YOLO model. This newly labeled dataset was employed for fine-tuning the STAR model directly on the edge training node. Following a single cycle of closed-loop adaptation, the updated model was re-deployed to the detection node for performance re-evaluation.

As shown in Table IV, the maxVSTAR adaptation process produced a dramatic recovery in classification accuracy, with the overall average increasing from 49.14% to 81.51%. Notably, activities that initially suffered severe degradation, such as "stand up" and "run," exhibited substantial improvements to 77.68% and 80.92%, respectively. This performance restoration validates the efficacy of the cross-modal supervisory mechanism and confirms that maxVSTAR can autonomously compensate for operational discrepancies introduced by hardware variations or environmental changes — all while preserving data privacy and avoiding reliance on cloud infrastructure.

TABLE IV. MODEL CLASSIFICATION ACCURACY AFTER MAXVSTAR RETRAINING

| Class of activity | Accuracy |
|---|---|
| lie down | 86.90% |
| fall | 69.41% |
| walk | 79.19% |
| pickup | 90.11% |
| run | 80.92% |
| sit down | 86.10% |
| stand up | 77.68% |
| Have a person or No person | 91.32% |

These experimental results substantiate the central hypothesis of this research. The sharp decline in HAR performance following domain shift confirms that environment-specific characteristics critically impair the transferability of static CSI-based models. More importantly, the successful recovery of recognition accuracy through a single iteration of vision-assisted fine-tuning validates the practical utility of the proposed maxVSTAR framework. By creating a dynamic, vision-guided adaptation loop that autonomously updates the "student" CSI model using a reliable "teacher," this system effectively addresses a longstanding challenge in device-free HAR research: the lack of robust, self-adaptive solutions suitable for privacy-sensitive, edge-deployed IoT environments.

To the best of our knowledge, no prior study has demonstrated an operationally validated, closed-loop, cross-modal HAR system capable of autonomously correcting for domain shifts entirely at the edge. The ability of maxVSTAR to mitigate hardware inconsistencies without cloud interaction or human intervention positions it as a highly practical and scalable solution for long-term deployment in smart homes, healthcare monitoring, and industrial IoT applications.

## VI. Conclusion

This study successfully proposed and validated maxVSTAR, an end-to-end, edge-deployable framework designed to address one of the most persistent challenges in WiFi Channel State Information (CSI)-based Human Activity Recognition (HAR): the problem of domain shift. While previous works have demonstrated promising recognition performance under controlled conditions, their static nature renders them highly sensitive to hardware variability and environmental dynamics, severely limiting their reliability in real-world, long-term deployments. In contrast, maxVSTAR introduces a dynamic, closed-loop, vision-assisted adaptation mechanism that enables continuous, autonomous model recalibration directly at the network edge, thereby mitigating these limitations without reliance on cloud-based computation or human intervention.

The core innovation of this work lies in the integration of a high-precision YOLO-based vision "teacher" system, enhanced through the introduction of the C2f_iRMB module, with a lightweight CSI-based "student" model. This cross-modal supervision strategy enables on-demand, real-time generation of reliable activity labels, which are then used to fine-tune the CSI model in situ. Through this design, the system not only preserves user privacy during standard operation—since visual data is utilized solely during temporary calibration phases—but also maintains resilience against hardware-induced discrepancies and environmental variability.

Experimental validation demonstrated both the severity of domain shift in conventional static CSI-HAR models and the effectiveness of the proposed solution. The baseline STAR model, which achieved an initial recognition accuracy of 93.52% under calibrated conditions, experienced a substantial degradation to 49.14% when deployed on uncalibrated hardware. Following a single iteration of the maxVSTAR closed-loop adaptation process, classification accuracy recovered to 81.51%, confirming the viability of the proposed framework as a practical and scalable solution for dynamic, privacy-sensitive IoT environments.

While the present work offers a significant advancement, several limitations remain. The evaluations were conducted within a constrained activity set and indoor environment, and the hardware-specific optimizations, while effective, may require adaptation for broader hardware platforms. Moreover, the framework currently operates in a cross-modal, sequential supervision paradigm rather than a fully integrated multi-modal fusion scheme. Addressing these limitations presents a clear direction for future research, particularly in exploring feature-level fusion strategies, unsupervised domain adaptation techniques, and efficient few-shot or continual learning algorithms that could further enhance the autonomy and generalizability of edge-deployed sensing systems.

By far this research establishes a novel and practical framework for overcoming domain shift in wireless HAR, moving the field beyond the limitations of static model deployment. It offers a concrete, deployable paradigm for realizing dynamic, robust, and self-adaptive perceptual systems, with broad applicability in smart home automation, ambient healthcare monitoring, and context-aware security infrastructure. The maxVSTAR system represents a foundational contribution toward the long-term vision of intelligent, edge-based sensing networks capable of maintaining operational integrity in complex, evolving environments without compromising data privacy or incurring prohibitive maintenance costs.

## Acknowledgment

This research was funded by the Guangzhou Development Zone Science and Technology Project (2023GH02), the University of Macau (MYRG2022-00271- FST) and research grant by the Science and Technology Development Fund of Macau (0032/2022/A).